\documentclass[copyright]{eptcs}
\providecommand{\event}{Y. Deville and C. Solnon (Eds): Sixth International Workshop\\
on Local Search Techniques in Constraint Satisfaction (LSCS'2009).
} % Name of the event you are submitting to
\providecommand{\volume}{5}
\providecommand{\anno}{2009}
\providecommand{\firstpage}{55}
\providecommand{\eid}{5}
\usepackage{breakurl}        % Not needed if you use pdflatex only.

\usepackage{amsmath}
\usepackage{amssymb}
\usepackage{latexsym,xspace,algorithm2e}
\usepackage{graphics,graphicx,float}
\usepackage{times,psfrag}
\usepackage{pstricks-add}
\usepackage{rotate}
\usepackage{epsfig}

\newcommand{\hybrid}{\textsc{SatHyS}\xspace}

\newcommand{\cls}{\textsc{cls}\xspace}
\newcommand{\wsat}{\textsc{wsat}\xspace}
\newcommand{\rsaps}{\textsc{rsaps}\xspace}
\newcommand{\adaptg}{\textsc{adaptg2}\xspace}
\newcommand{\minisat}{\textsc{minisat}\xspace}
\newcommand{\hybridGM}{\textsc{hybridGM}\xspace}
\newcommand{\hybridWei}{\textsc{hybrid1}\xspace}

\renewcommand{\overline}[1]{\bar{#1}}

\title{
  Integrating Conflict Driven Clause Learning to Local Search
}
\author{
Gilles Audemard~~~~~~~~~Jean-Marie Lagniez~~~~~~~~~Bertrand Mazure~~~~~~~~~Lakhdar Sa\"is
\thanks{supported by \textsc{anr unloc} project \textsc{anr08-blan-0289-01}}
\institute{Universit\'e Lille-Nord de France\\
CRIL - CNRS UMR 8188\\
Artois, F-62307 Lens\\
}
  \email{\{audemard,lagniez,mazure,sais\}@cril.fr}
}

\def\titlerunning{ Integrating Conflict Driven Clause Learning to Local Search
}
\def\authorrunning{G. Audemard \emph{et al.}}

\floatstyle{plain} %ruled plain, boxed
\newfloat{Algorithm}{thp}{lop}
\floatname{Algorithm}{Algorithm}

\newtheorem{definition}{Definition}
\newtheorem{example}{Example}
\newtheorem{proposition}{Proposition}

\begin{document}

% \title{%
%   Integrating Conflict Driven Clause Learning to Local Search
%   }

% \author{Gilles Audemard \and Jean-Marie Lagniez \and Bertrand Mazure \and Lakhdar Sais
% \thanks{supported by \textsc{anr unloc} project \textsc{anr08-blan-0289-01}}
% }
% \institute{CNRS, UMR 8188, F-62300 Lens, France\\
%   \{audemard,lagniez,mazure,sais\}@cril.univ-artois.fr
% }

\maketitle

\begin{abstract}
  This article introduces \hybrid ({\em SAT HYbrid Solver}), a novel
  hybrid approach for propositional satisfiability. It combines local
  search and conflict driven clause learning (CDCL) scheme. Each time
  the local search part reaches a local minimum, the CDCL is launched.
  For SAT problems it behaves like a tabu list, whereas for UNSAT
  ones, the CDCL part tries to focus on minimum unsatisfiable
  sub-formula (MUS). Experimental results show good performances on
  many classes of SAT instances from the last SAT competitions.
\end{abstract}

\section{Introduction}
The SAT problem, namely the issue of checking whether a set of Boolean
clauses is satisfiable or not, is a central issue in many computer
science and artificial intelligence domains, like  theorem
proving, planning, non-monotonic reasoning, VLSI correctness
checking. These last two decades, many approaches have been proposed
to solve large SAT instances, based on logically complete or
incomplete algorithms. Both local-search techniques
\cite{SelmanKC94,SelmanK93,HirschK05} and elaborate variants of
the Davis-Putnam-Loveland-Logemann DPLL procedure \cite{DPLL62}
\cite{MoskewiczMZZM01,EenS03}, called modern SAT solvers, can
now solve many families of hard SAT instances. These two kinds of
approaches present complementary features and performances. Modern SAT
solvers are particularly efficient on the industrial SAT category
while local search performs better on random SAT instances.

Consequently, combining stochastic local search (SLS) and conflict
driven clause learning (CDCL) solvers seems promising. Note that it
was pointed as a challenge by Selman {\it et al.} \cite{SelmanKM97} in
1997. Such methods should exploit the quality and differences of both
approaches. Furthermore, the perfect hybrid method has to outperform
both local search and CDCL solvers. A lot of attempts have been done
last decade \cite{HandbookOfSAT2009}. These different attempts will be
discussed in section \ref{sec:related}.

In this paper, we propose another hybridization of local search and
modern SAT solver, named \hybrid ({\em SAT HYbrid Solver}).  The local
search solver is the main one. Each time it reaches a local minimum,
the CDCL part is called and assigns some variables.  This part of our
solver is expected to have different behaviours depending on the kind
of formula to solve. In case of a satisfiable one, the CDCL part can
be seen as a tabu list \cite{GloverI89} in order to protect good
variables and avoid to reach the same minimum quickly.  On the other
hand, for unsatisfiable formulas, it tries to focus the search on
minimum unsatisfiable sub-formulas (MUS)
\cite{GMP-06-2,GMP-07-1,GMP-07-2}, allowing to concentrate on a small
part of the whole formula.  Like this, the CDCL component of \hybrid
is used as a strategy to escape from local minimum.

The rest of the paper is organized as follows.  Section 2 introduces
different notions necessary for understanding the rest of the
paper. Section 3 discusses different hybrid methods.  Section 4 gives
the insights of our method. In section 5, we give the details and
algorithms of \hybrid.  Before a conclusion, section 6 provides
different experiments.

\section{Preliminary definitions and technical background}
\label{sec:predef}
\subsection{Definitions}

Let us give some necessary definitions and notations.  Let \(
V=\{x_{1}...x_{n}\} \) be a set of boolean variables, a literal \(
\ell \) is a variable $x_{i}$ or its negation $\overline{x_{i}}$. A
clause is a disjunction of literals $c_{i}=(\ell_{1}\vee
\ell_{2}...\vee \ell_{n_{i}})$.  A unit clause is a clause with only
one literal. A formula $\Sigma$ is in conjunctive normal form (CNF) if
it is a conjunction of clauses $\Sigma=(c_{1}\wedge c_{2}...\wedge
c_{m})$.  The set of literals appearing in $\Sigma$ is denoted
$\mathcal{V}_{\Sigma}$. An interpretation $\mathcal{I}$ of a formula
$\Sigma$ associates a value $\mathcal{I}(x)$ to variables in the
formula. An interpretation is {\it complete} if it gives a value to
each variable $x\in \mathcal{V}_{\Sigma}$, otherwise it is said {\it
  partial}. A clause, a CNF formula and an interpretation can be
conveniently represented as sets. A {\it model} of a formula $\Sigma$,
denoted $\mathcal{I} \models \Sigma$, is an interpretation
$\mathcal{I}$ which satisfies the formula $\Sigma$ i.e. satisfies each
clause of $\Sigma$. Then, we can define the SAT decision problem as
follows: is there an assignment of values to the variables so that the
CNF formula $\Sigma$ is satisfied?

\noindent Let us introduce some additional notations.

\begin{itemize}
\item The negation of a formula $\Gamma$ is denoted $\overline{\Gamma}$
\item $\Sigma_{|_\ell}$ denotes the formula $\Sigma$ simplified by the
  assignment of the literal $\ell$  to true. This notation is extended
  to interpretations: Let  $\mathcal{P} = \{\ell_1,...,\ell_n\}$ be an
  interpretation,      $\Sigma_{|_{\mathcal{P}}}     =     (...(\Sigma
  |_{\ell_1})...|_{\ell_n})$ ;
\item  $\Sigma^*$  denotes the  formula  $\Sigma$  simplified by  unit
  propagation;
\item $\models_*$ denotes logic deduction by unit propagation: $\Sigma
  \models_*  l$  means  that  the  literal $x$  is  deducted  by  unit
  propagation       from      $\Sigma$       i.e.       $(\Sigma\wedge
  \overline{\ell})^*=\bot$ . One notes  $\Sigma \models_* \bot$ if the
  formula is unsatisfiable by unit propagation.
\item $\eta[x, c_i, c_j]$ denotes the \emph{resolvent} between a
  clause $c_i$ containing the literal $x$ and $c_j$ a clause
  containing the opposite literal $\neg x$.  In other words $\eta[x,
  c_i, c_j] = c_i\cup c_j\backslash \{x, \neg x\}$.
  % a  {\it  resolvent} on  $x$  of $c_i$  and  $c_j$,  is defined  as
  % $\eta[x, c_i, c_j] = c_i\cup c_j\backslash \{x, \neg x\}$.
  A resolvent  is called {\it tautological} when  it contains opposite
  literals.
\end{itemize}

\subsection{\label{sec:wsat}Local Search Algorithms}

Local search  algorithms for SAT  problems use a stochastic  walk over
complete  interpretations of  $\Sigma$. At  each {\em  step}  (or {\em
  flip}),  they try  to  reduce the  number  of unsatisfiable  clauses
(usually  called  a descent).   The  next  complete interpretation  is
chosen among  the neighbours of the  current one (they  differ only on
one literal  value).  A  local minimum is  reached when no  descent is
possible. One of  the key point of stochastic  local search algorithms
is the method  used to escape from local minimum.   For lack of space,
we  can  not provide  a  general  algorithm  of local  search  solver.
%However, a  modified version can be seen  in algorithm \ref{alg:CDLS}.
For more details, the reader will refer to \cite{hoos04}.

\subsection{CDCL solvers}

\restylealgo{ruled}\linesnumbered
\begin{algorithm}[t]
  \caption{\label{alg:cdcl}CDCL solver}
\SetLine
%\SetVline
\SetInd{0.3em}{0.7em}
\KwIn{a CNF formula $\Sigma$}
\KwOut{SAT or UNSAT}
{
  $I = \emptyset$ \tcc*{interpretation} 
  $dl = 0$ \tcc*{decision level}
  $x_c = 0$ \tcc*{number of conflicts}
  \While{(true)}{
    $\gamma=$ BCP($\Sigma$,$I$)\;
    \uIf{($\gamma$!=null)}{
      $x_c=x_c+1$\;
      $\beta=$conflictAnalysis($\Sigma$,$I$,$c$)\;
      $bl =$ computeBackjumpingLevel($\gamma$,$I$)\; 
      \lIf{($bl<0$)}{\textbf{return} UNSAT\;}
      $\Sigma=\Sigma\cup \{\gamma\}$\;
      \lIf{(restart())}{$bl=0$\;}
      backjump($\Sigma$,$I$,$bl$)\;
      $dl = bl$\;
    }\uElse
    {
      \uIf{(all variables are instanciated)}
      {\textbf{return} SAT\;}
      $\ell$ = chooseDecisionLiteral($\Sigma$)\;
      $dl = dl+1$\;
      $I = I\cup\{\ell\}$\;
    }
  }
}

\end{algorithm}

Algorithm  \ref{alg:cdcl} shows the  general scheme  of a  CDCL solver
(due  to  lack   of  space,  we  can  not   provide  details  for  all
subroutines).  A typical  branch of  a CDCL  solver is  a  sequence of
decisions,  followed by  propagations,  repeated until  a conflict  is
reached. Each decision  literal (lines 18--20) is assigned  at a given
decision level  ($dl$), deducted  literals (by unit  propagation) have
the  same  decision  level.   If  all  variables  are  assigned,  then
$\mathcal{I}$  is a  model of  $\Sigma$  (lines 16--17).  Each time  a
conflict is reached by unit propagation (then $\gamma$ is the conflict
clause) A  nogood $\beta$ is computed  (line 8) using  a given scheme,
usually the first-UIP (Unique Implication Point) one \cite{ZhangMMM01}
and a  backjump level is computed.  At this point, It may have proved
the unsatisfiability of  the formula $\Sigma$. If it  is not the case,
the nogood  $\beta$ is  added to the  clause database and  backjump is
done (lines 11--14).  Finally, sometimes CDCL solvers enforce restarts
(different strategies are possible  \cite{Huang07}). In this case, one
backjump in the top of the search tree.

\subsection{Muses}

Minimum unsatisfiable sub-formulas (MUS) of a CNF formula represent the
smallest explanations for the inconsistency in term of the number of
clauses. MUS are very important in order to circumscribe and
highlight the source of contradiction of a given formula. Formally, one has:

\begin{definition}
Let $\Sigma$ be a CNF formula. A MUS $\Gamma$ of $\Sigma$ is a set of clauses such that:
\begin{enumerate}
\item $\Gamma \subseteq \Sigma$;
\item $\Gamma$ is unsatisfiable;
\item $\forall \Delta \subset \Gamma, \Delta$ is satisfiable.
\end{enumerate}
\end{definition}

\begin{figure}
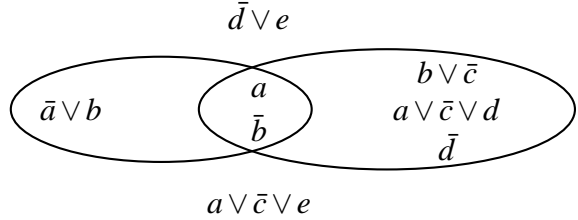

  \begin{center}
    \large{
      \pspicture(10,3.5)

      \psellipse(3.7,1.75)(2,0.7)
      \psellipse(6.7,1.75)(2.5,0.8)
      
      \rput(5,3){$\overline{d} \vee e$}
      
      \rput(5,2){$a$}
      \rput(5,1.5){$\overline{b}$}
      
      \rput(2.5,1.75){$\overline{a} \vee b$}
      
      \rput(7.5,2.25){$ b \vee \overline{c}$}
      \rput(7.5,1.75){$ a \vee \overline{c} \vee d$}
      \rput(7.5,1.25){$\overline{d}$}
      
      \rput(5,0.5){$a \vee \overline{c} \vee e$} 
      
      \endpspicture
    }
  \end{center}
  \caption{All MUS of the formula $\Sigma$ (example \ref{fig:exmus})}
  \label{fig:MUS}

\end{figure}

\begin{example}
  \label{fig:exmus}
  Let $\Sigma = \{\overline{d} \vee e,~ b \vee \overline{c},~
  \overline{d},~ \overline{a} \vee b,~ a,~ a \vee \overline{c} \vee e
  ,~\overline{a} \vee c \vee d,~ \overline{b}\}$ be a CNF formula.
  Figure \ref{fig:MUS} represents all MUS of $\Sigma$.\end{example}

Due to unsatisfiability of MUS, one has the following property:

\begin{proposition}
  Let  $\Sigma$ be  an unsatisfiable  CNF formula,  $\Gamma$ a  MUS of
  $\Sigma$. 
  \begin{center}
    $\forall \mathcal{I}$ an interpretation over $\mathcal{V}_\Sigma$, $\exists c \in \Gamma$ such that $\mathcal{I} \not \models c$
  \end{center}
\end{proposition}

Let us consider  a CNF formula $\Sigma$ and  a complete interpretation
$\mathcal{I}_c$.   We   say   that   the  literal   $\ell$   satisfies
(resp. falsifies) a  clause $\beta\in\Sigma$ if $\ell\in \mathcal{I}_c
\cap \beta$ (resp.  $\ell\in \mathcal{I}_c \cap \overline{\beta}$). We
note           $\mathcal{L}^+_{\mathcal{I}_c}(\beta)$           (resp.
$\mathcal{L}^-_{\mathcal{I}_c}(\beta)$),    the   set    of   literals
satisfying  (resp.   falsifying)  a  clause   $\beta$.  The  following
definitions were introduced in \cite{Gregoire-etal:2006a}.

\begin{definition}[once-satisfied clause]
  A clause $\beta$ is said {\em once-satisfied} by an interpretation
  $\mathcal{I}_c$ on literal $z$ if $\mathcal{L}^+_{\mathcal{I}_c}(\beta) = \{z\}$.
\end{definition}

\begin{definition}[critical and linked clauses]
  Let $\mathcal{I}_c$ be a  complete interpretation. A clause $\alpha$
  is          critical         wrt          $\mathcal{I}_c$         if
  $|\mathcal{L}^+_{\mathcal{I}_c}(\alpha)| = 0$  and $\forall \ell \in
  \alpha$,  $\exists  \alpha' \in  \Sigma$  with $\overline{\ell}  \in
  \alpha'$  and  $|\mathcal{L}^+_{\mathcal{I}_c}(\alpha')|=1$. Clauses
  $\alpha'$  are  {\it  linked}  to $\alpha$  for  the  interpretation
  $\mathcal{I}_c$.
\end{definition}

\begin{example}
  Let      $\Sigma=(\overline{a}      \vee      \overline{b}      \vee
  \overline{c})\wedge(a      \vee      \overline{b})\wedge(b      \vee
  \overline{c})\wedge(c   \vee  \overline{a})$   be   a  formula   and
  $\mathcal{I}_c=\{a,b,c\}$ an interpretation.  The clause $\alpha_1 =
  (\overline{a} \vee \overline{b} \vee \overline{c})$ is critical. The
  other   clauses   of  $\Sigma$   are   linked   to  $\alpha_1$   for
  $\mathcal{I}_c$.
\end{example}

The  following  properties was  proposed  and  exploited  in order  to
compute MUS by \cite{Gregoire-etal:2006a}.

\begin{proposition}
  \label{prop:1}
  In a  minimum (local  or global), the  set of falsified  clauses are
  critical.
\end{proposition}

\begin{proposition}\label{prop:criticalMUS}
  \label{prop:2}
  In a minimum (local or global), at least one of clause of each MUS
  is critical.
\end{proposition}

\section{Related Works}
\label{sec:related}
As it was suggested in the introduction, a lot of different approaches
have been proposed to combine local search and DPLL based ones. One
can divide such hybridizations in three different categories depending
on the kind of the main solver.  First, the main solver can be the SLS
one. In that case, DP is used in order to help SLS
\cite{MazureSG98,Crawford93,Ferris2004,Havens2004}. All of these
approaches use the local search component as an assistance for the
heuristic choice for variable assignment. Some of them try to focus
the search on the unsatisfiable part of the formula \cite{MazureSG98},
others on the satisfiable one \cite{hybrideGM2009,Ferris2004}. Furthermore, this step can be
achieved before the search \cite{Crawford93} or dynamically at each
decision nodes \cite{MazureSG98}.

The second category of hybridizations is the opposite, that is, the
SLS solver is the core of the method and the DPLL one helps it
\cite{HabetLDV02,Jussien2002}. In \cite{HabetLDV02}, the DPLL solver
is used in order to find dependencies between variables. Then, the
local search framework is called on a subset of variables (the
independent ones). Whereas, in \cite{Jussien2002} (note that this method is
for constraint satisfaction problems), the local search engine is used
to find a promising partial interpretation. 

Finally, the last category contains hybrid solvers where the both engines
work together \cite{FangH07,Letombe08}. The second method is an
improvement of the first one. The local search tries to find a
solution. After some time, it stops and sends all falsified
clauses by the current interpretation to the CDCL part. This last one
has the responsibility to find a model to this sub-formula. If it
proves unsatisfiable, then the whole instance is unsatisfiable too.

\begin{figure}
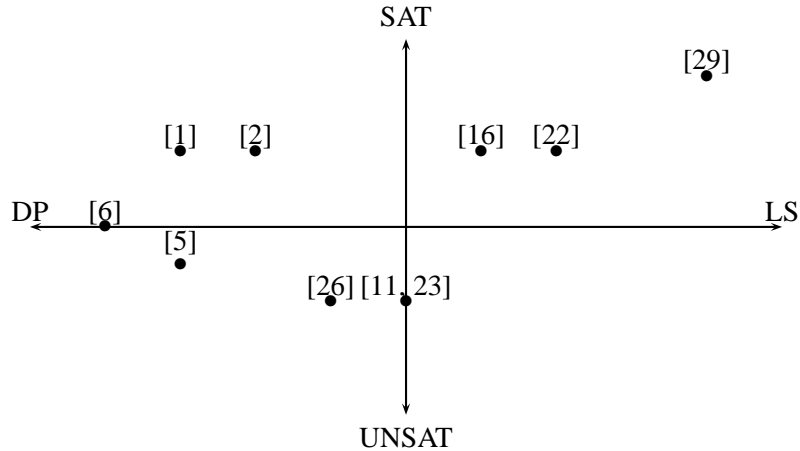

  \begin{center}
    \pspicture(10, 5)
    
    \psline{<->}(0,2.5)(10,2.5)
    \psline{<->}(5,0)(5,5)    
    
    \rput(0,2.7){DP}
    \rput(10,2.7){LS}
    \rput(5,5.3){SAT}
    \rput(5,-0.3){UNSAT}

    \rput(1,2.5){$\bullet$}
    \rput(1,2.7){\cite{DPLL62}}

    \rput(9,4.5){$\bullet$}
    \rput(9,4.7){\cite{SelmanKC94}}

    \rput(2,2){$\bullet$}
    \rput(2,2.3){\cite{Crawford93}}

    \rput(2,3.5){$\bullet$}
    \rput(2,3.7){\cite{Ferris2004}}

    \rput(3,3.5){$\bullet$}
    \rput(3,3.7){\cite{hybrideGM2009}}

    \rput(6,3.5){$\bullet$}
    \rput(6,3.7){\cite{HabetLDV02}}

    \rput(7,3.5){$\bullet$}
    \rput(7,3.7){\cite{Jussien2002}}

    \rput(4,1.5){$\bullet$}
    \rput(4,1.7){\cite{MazureSG98}}

    \rput(5,1.5){$\bullet$}
    \rput(5,1.7){\cite{FangH07,Letombe08}}

    \endpspicture  
  \end{center}
  \caption{Classification}
  \label{fig:class}
\end{figure}

We propose in Figure \ref{fig:class} a classification of all of these
approaches. The X-axis corresponds to the kind of search. For example,
DPLL is at the left, whereas walksat is at the right. The Y-axis
corresponds to the ability to solve SAT and/or UNSAT formulas. Then,
walksat is at the top of the classification.  Methods introduced above
are located in this graph. Of course, this classification is
subjective and and it can be subject of discussion. It is here to help
the reader to understand all of these approaches.

\section{Intuition}

In this section, we provide insights of our hybrid approach
\hybrid. They are related to the satisfiability or unsatisfiability of
the formulas. First note that the SLS engine is the core of our
method. Then, the Local search part tries to find a solution. When a
local minimum is reached, the CDCL part of the solver is launched. It
works like a tabu list in case of satisfiable formula and tries to
focus on MUS for unsatisfiable ones. Let us explain the main
differences now.

\subsection{SAT instances}

Much  research has  been done  on meta-heuristics.   Among  them, Tabu
search was introduced in  1986 by Glover \cite{GloverI89} and extended
to the SAT case in 1995  \cite{MSG-95-1}. The main idea of tabu search
consists,   in  a  given   position  (interpretation),   in  exploring
neighbours and choosing  as the next position the  one which minimises
the objective function.

It  is crucial  to  note that  such  an operation  could increase  the
objective function value: it is the case when  all neighbours have a
greater  value.  Then,  this  mechanism allows  to  escape from  local
minimum.  However,  the main  drawback is that  at the next  step, one
goes back  in the same local  minimum. To avoid  this, heuristic needs
memory  for  the  last  explored  positions  to  be  forbidden.  These
positions are \textit{tabu}.

Already  explored positions  are  stored in  a  queue (usually  called
\textit{tabu  list}) of a  given length  which is  a parameter  of the
method.  This  list must  contain  complete  positions,  which can  be
prohibitive. To  go round this,  one can store only  previous actions,
associated to values of  the objective function.  The length parameter
is very  important. A lot  of work have  been done to  provide optimal
length,      statically      \cite{MazureSG98}     or      dynamically
\cite{Battiti1997}.

We propose to keep the set of tabu variables by using a partial
interpretation computed with unit propagation engine. When a variable
becomes tabu, it is assigned in the CDCL solver part and propagated.
Then, resulting interpretation is used as a tabu list. There are two
advantages: firstly, the length of the tabu list is dynamic, it depends
of unit propagation and backjumping. Secondly, unit propagation allows
to catch some functional dependencies in the tabu list.

\subsection{UNSAT instances}

First of all, note that if an instance is unsatisfiable then, whatever
is    the     complete    interpretation,    a     falsified    clause
exists. Furthermore, if an instance is unsatisfiable, then it contains
at least one  MUS. This MUS, i.e. a subset of  clauses of the formula,
is often smaller  than the global formula and,  then, can contain less
variables.   Then,  in  the  case  of  unsatisfiable  formula,  it  is
advantageous to focus the search on such variables.

In    the    frame   of    MUS    detection,    Gr\'egoire   {\it et    al.}
\cite{Gregoire-etal:2006a,GMP-07-2}  shown that local  search provides
good heuristics, concerning  inconsistent kernel detection. These methods
use properties \ref{prop:1} and \ref{prop:2}
in order to balance clauses which could be part of a MUS.

The proposed method  in this paper is  based on this principle.  When a local
minimum  is reached,  property  \ref{prop:1} assures  that  the set  of
clauses falsified  by current interpretation are  critical. Given that
such clauses could be part of a  MUS, we choose one of them to make it
totally true. Therefore only the variables of a kernel are expected to
be taken into account.

\section{Implementation}

As explicated in the previous  section, the core of our solver \hybrid
is  the local search  component. It  is based  on an  iterative search
process that in  each step moves from one point  to a neighbouring one
until discovering  a solution.   At each step  it tries to  reduce the
number of falsified clauses. When  it is not possible, a local minimum
is  reached. In  that case,  the CDCL  part is  called.  It  chooses a
falsified clause and assigns all  of its literals such that the clause
becomes totally valid. All literals  of the chosen clause are decision
nodes.  Of  course unit  propagation is achieved.  In this  manner, it
escapes from the  local minimum and the SLS part  of the hybrid solver
can be used again.  Note that  all variables assigned by the CDCL part
are fixed and can not be flipped by the SLS solver.  Of course, during
the  CDCL  process, a  conflict  can  occur.  In that  case,  conflict
analysis  is performed, a  clause is  learnt and  a backjump  is done.
Then, some  of fixed variables become  free and can  be flipped again.
This   conflict   analysis   makes    the   solver   able   to   prove
unsatisfiability.

\restylealgo{ruled}\linesnumbered
\begin{algorithm}[t]
\caption{\hybrid}
\label{alg:hybride}
\SetLine
\SetVline
\SetInd{0.3em}{0.7em}
\KwIn{$\Sigma$ a CNF formula}
\KwResult{$SAT$ if $\Sigma$ is satisfiable, else $UNSAT$}
{
  \While{($true$)}
  {
    $\mathcal{I}_c \leftarrow$ Init($\Sigma$)\;
    $\mathcal{I}_p \leftarrow \emptyset$\; 
    \For{$j \leftarrow 1$ to $MaxFlips$}
    {
      \If{$\mathcal{I}_c \models \Sigma_{|\mathcal{I}_p}$}
      {
        \Return $SAT$\;
      }
       $\Gamma = \{\alpha \in \Sigma_{|\mathcal{I}_p} |~\mathcal{I}_c \not \models \alpha\}$\tcc*[f]{set of falsified clauses}\;
      \While{$\Gamma \not = \emptyset$}
      {
        $\alpha \in \Gamma$\;
        \eIf{$\exists x \in \alpha$ allowing a descent}
        {
          $flip(x)$\;
          break\;
        }
        {
          $\Gamma \leftarrow \Gamma \setminus \{\alpha\}$\;
        }       
      }
      \If(\tcc*[f]{local minimum}){$\Gamma = \emptyset$}
      {
        $\alpha \in \Sigma_{|\mathcal{I}_p}$ such that $\mathcal{I}_c \not \models \alpha$\;
        \If{($fix(\Sigma, \mathcal{I}_c, \mathcal{I}_p, \alpha)$=$UNSAT$)}
        {
          \Return UNSAT\;
        }
      }
    }
  }
}
\end{algorithm}

Algorithm \ref{alg:hybride} takes a  CNF formula $\Sigma$ in parameter
and returns  SAT or UNSAT.  It is based  on WSAT-like  algorithms. Two
variables are used. A  complete interpretation $\mathcal{I}_c$ for the
local   search   engine   (initialised   randomly)   and   a   partial
interpretation $\mathcal{I}_p$  for the CDCL part  (initialised to the
empty set).   In order to  forbid to flip  fixed literals by  the CDCL
part  (the literals  of $\mathcal{I}_p$),  the SLS  solver  deals with
$\Sigma_{|\mathcal{I}_p}$. If the current complete interpretation is a
model of  $\Sigma_{|\mathcal{I}_p}$ then \hybrid  finishes and returns
SAT   (lines  5--6).   Otherwise,  if   it  exists   a   neighbour  of
$\mathcal{I}_c$  which  allows to  decrease  the  number of  falsified
clauses,  it  becomes   the  current  complete  interpretation  (lines
8--14). If it  is not the case, then a local  minimum is reached (line
15).  In that  case, a  falsified clause  is randomly  chosen  and the
function  {\tt fix}  is called  in order  to fix  new  literals (lines
15--17). This function is explained below. It modifies interpretations
$\mathcal{I}_c$ and $\mathcal{I}_p$ by  fixing new variables and (if a
conflict  occurs during  boolean propagation)  freeing other  ones. At
this step, the CDCL solver  can prove the unsatisfiability. Of course,
if it is the case the search is done (line 17--18).

This whole  process is  repeated a given  number of  times ($MaxFlips$,
line 4).  After that, the  solver tries to  go in another area  of the
search space. Then, the process can continue until finding an answer.

Function {\tt fix} is  described in Algorithm \ref{alg:fixe}. It works
like  a  very  simple CDCL  solver.  It  takes  a clause  $\alpha$  in
input.   It  takes   also   in  input   the  complete   interpretation
$\mathcal{I}_c$  and  the  partial  one $\mathcal{I}_p$  and  modifies
them.  It  returns $UNSAT$  if  the  unsatisfiability  is proven,  and
$UNKNOWN$  otherwise. The main  goal of  this function  is to  fix new
variables. To  achieve this,  it tries to  totally satisfy  the clause
$\alpha$. First of  all, the set of decision  denoted $\mathcal{E}$ is
initialized.  Whenever it is not  empty and a conflict does not occur,
a  new  decision   variable  is  chosen  and  added   to  the  partial
interpretation and boolean unit  propagation (BCP) is performed (lines
3--6).  If a conflict occurs,  then the process is stopped. A conflict
analysis  is   done  and   the  partial  interpretation   is  repaired
(\texttt{backjumping}).   At  this step  the  unsatisfiability can  be
proved. Otherwise, the obtained nogood  is added to the clause database
(lines 7--11).   Then, the complete  interpretation $\mathcal{I}_c$ is
updated with  the help of  the partial one (note  that $\mathcal{I}_p$
and $\mathcal{I}_c$ can not differ).

\restylealgo{ruled}\linesnumbered
\begin{algorithm}[t]
\caption{fix}
\label{alg:fixe}
\SetLine \SetVline \SetInd{0.3em}{0.7em} \KwIn{$\alpha$ a clause}
\KwOut{$\Sigma$ a CNF,
  $\mathcal{I}_c$ a complete interpretation,
  $\mathcal{I}_p$ a partial interpretation}
\KwResult{$UNSAT$ if unsatisfiable is proven, $UNKNOWN$ otherwise}

\linesnumbered { 
  $\gamma \leftarrow \emptyset$\;
  $\mathcal{E} \leftarrow \{ x |~ \overline{x} \in \alpha\}$\;
  \While{$(\mathcal{E} \not = \emptyset)$ and $(\alpha = \emptyset)$}
  {
    $\mathcal{I}_p \leftarrow \mathcal{I}_p \cup \{x\}$ tel que $x \in \mathcal{E}$\;
    $\gamma \leftarrow BCP()$\;
    $\mathcal{E}\leftarrow\mathcal{E}\setminus\{x\in\mathcal{E}|~x\in\mathcal{I}_p~or~\overline{x}\in\mathcal{I}_p\}$\;
  }
  \If{$\gamma \not = \emptyset$}
  {
    $\beta=$conflictAnalysis($\Sigma$,$\mathcal{I}_p$,$\gamma$)\;
    $bl =$computeBackjumpingLevel($\gamma$,$\mathcal{I}_p$)\; 
    \lIf{($bl<0$)}{\Return $UNSAT$\;}
    $\Sigma \leftarrow \Sigma \cup \{\beta\}$\;
  }
  $\rho \leftarrow \{x \in \mathcal{I}_c|~\overline{x} \in \mathcal{I}_p\}$\;
  $\mathcal{I}_c \leftarrow \mathcal{I}_c \setminus \{\overline{x}|~x\in\rho\} \cup \rho$\;
  \Return $UNKONWN$\;
}
\end{algorithm}

\section{Experiments}

Experimental results reported in this  section were obtained on a Xeon
3.2 GHz with 2 GByte of RAM. The CPU time is limited to 1200 seconds.

Our approach is compared with: 
\begin{itemize}
\item three SLS methods: 
  \begin{enumerate}
  \item classical \wsat \cite{SelmanKC94}, i.e. using random walk strategy
  \item \rsaps \cite{HutterTH02}
  \item \adaptg \cite{LiWZ07}
  \end{enumerate}
\item two recent hybrid methods submitted at the last SAT competition in 2009:
  \begin{enumerate}
  \item \hybridGM \cite{hybrideGM2009}
  \item \hybridWei \cite{LiWZ07}
  \end{enumerate}
\item and two complete methods:
  \begin{enumerate}
  \item \cls a local search method completed by adding resolution process \cite{Fang04}
  \item \minisat \cite{EenS03} a well-known CDCL solver.
  \end{enumerate}
\end{itemize}

Instances used are taken from the last SAT competitions
(\url{www.satcompetition.org}).  They are divided into three different
categories: crafted (1439 instances), industrial (1305) and random
(2172). All instances are preprocessed with SatElite \cite{biere05}.

\begin{table}[t]
  \centering
  \begin{tabular}{l|rr|rr|rr}
    & \multicolumn{2}{c|}{Crafted} & \multicolumn{2}{c|}{Industrial} & \multicolumn{2}{c}{Random} \\
    \cline{2-7}
    & sat & unsat & sat & unsat & sat & unsat \\
    \hline
    \adaptg & 326 & 0 & 232 & 0 & 1111 & 0 \\
    \rsaps&339 &0&226 & 0 & 1071 & 0\\
    \wsat&259 &0&206 & 0 & 1012 & 0\\
    \hline
    \cls&235 &75&227 & 102&690 & 0 \\
    \hline
    \hybrid &322& 191&466 &309 & 341 & 14\\
    \hybridGM  &290 & 0& 209 &0 & 1114 & 0\\
    \hybridWei  &329 & 0& 277 &0 & 1126 & 0\\
    \hline
    \minisat &402 & 369&588&414 & 609 & 315\\
  \end{tabular}
  \caption{  
    \label{tab:result}
    \hybrid versus some other SAT solvers }
\end{table}

\noindent Table \ref{tab:result} summarizes the  obtained results on this large
number of instances.  For more  details on this experimental part, the
reader  can refer  to  \texttt{http://www.cril.fr/$\sim$lagniez/sathys}.  For
each  category and  for each  solver we  report the  number  of solved
instances. Of course, \minisat  a state-of-the-art CDCL based complete
solver  is only considered  to  mention the  gap between  local
search based techniques and  complete modern SAT solvers on industrial
and crafted  instances. On random satisfiable  instances, local search
techniques generally outperform complete techniques.

Before analysing more precisely the table of results (Table
\ref{tab:result}), remark that only three solvers are able to solve
unsatisfiable instances (\minisat, \cls and \hybrid). The recent
hybrid methods submitted at the last SAT competition cannot prove 
inconsistency in the allowed time.

On the crafted instances, \hybrid is very competitive and solves
approximately the same number of satisfiable instances as \rsaps,
\adaptg and the recent hybrid methods. Furthermore, \hybrid solves
much more instances than \wsat, its built-in solver. Concerning
unsatisfiable crafted instances, as expected our approach is less
efficient than \minisat but it is proved highly more efficient than
\cls.

Concerning industrial instances, \hybrid solves two times more
satisfiable instances than SLS and hybrid methods. Once again, on
unsatisfiable industrial instances, your solver is better than \cls
but less efficient than \minisat.

These results show that conflict analysis allows to solve efficiently
structured SAT and UNSAT instances.

%%%%%%%%%%%%%%%%%%%%%%%%%%%%%%%%%%%%%%%

Finally, for the random category, we can note that \hybrid is unable
to solve unsatisfiable problems.  As pointed by \minisat results,
learning is not the good approach to solve random instances.
As a summary, unfortunately our approach cannot reach the minisat
performance. However the solver \hybrid is much more efficient than
local search based algorithms and hybrid methods. It significantly
improves \wsat, its built-in solver.
Even if \minisat is the best solver on crafted and industrial
instances, these first results are very encouraging and reduce the gap
between local search based techniques and DPLL-like complete solvers.

\begin{figure}
      \centerline{\epsfig{file=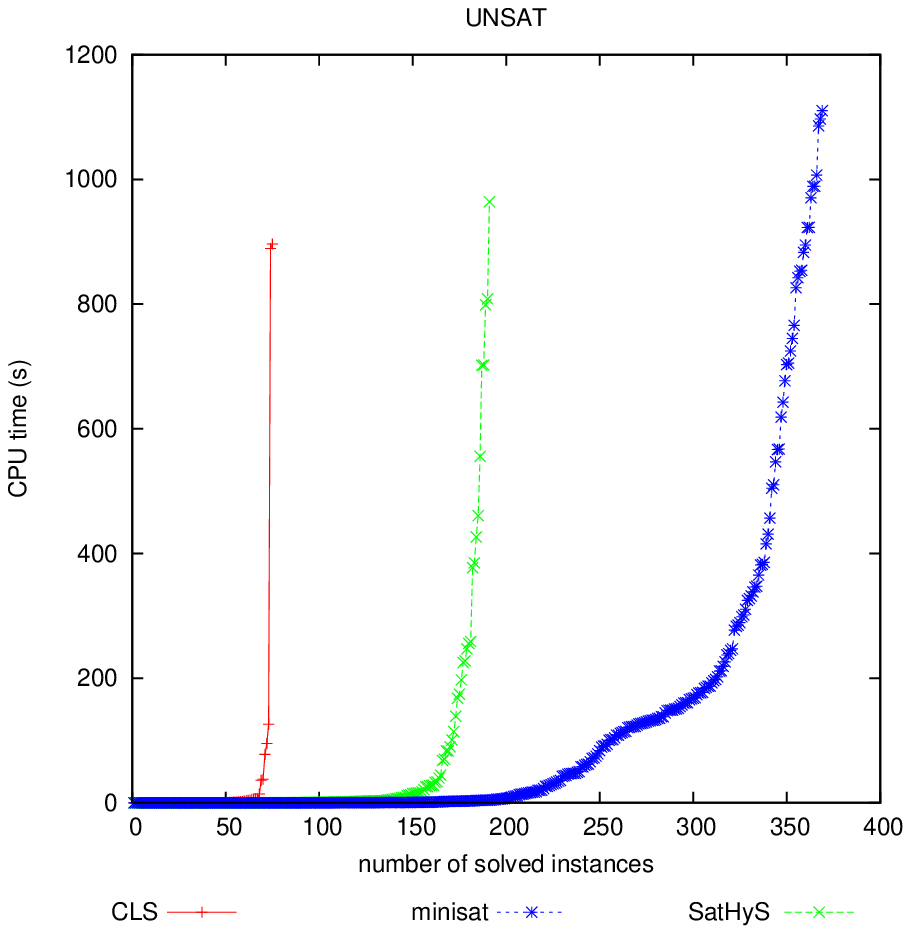,width=270pt}}
      \centerline{\epsfig{file=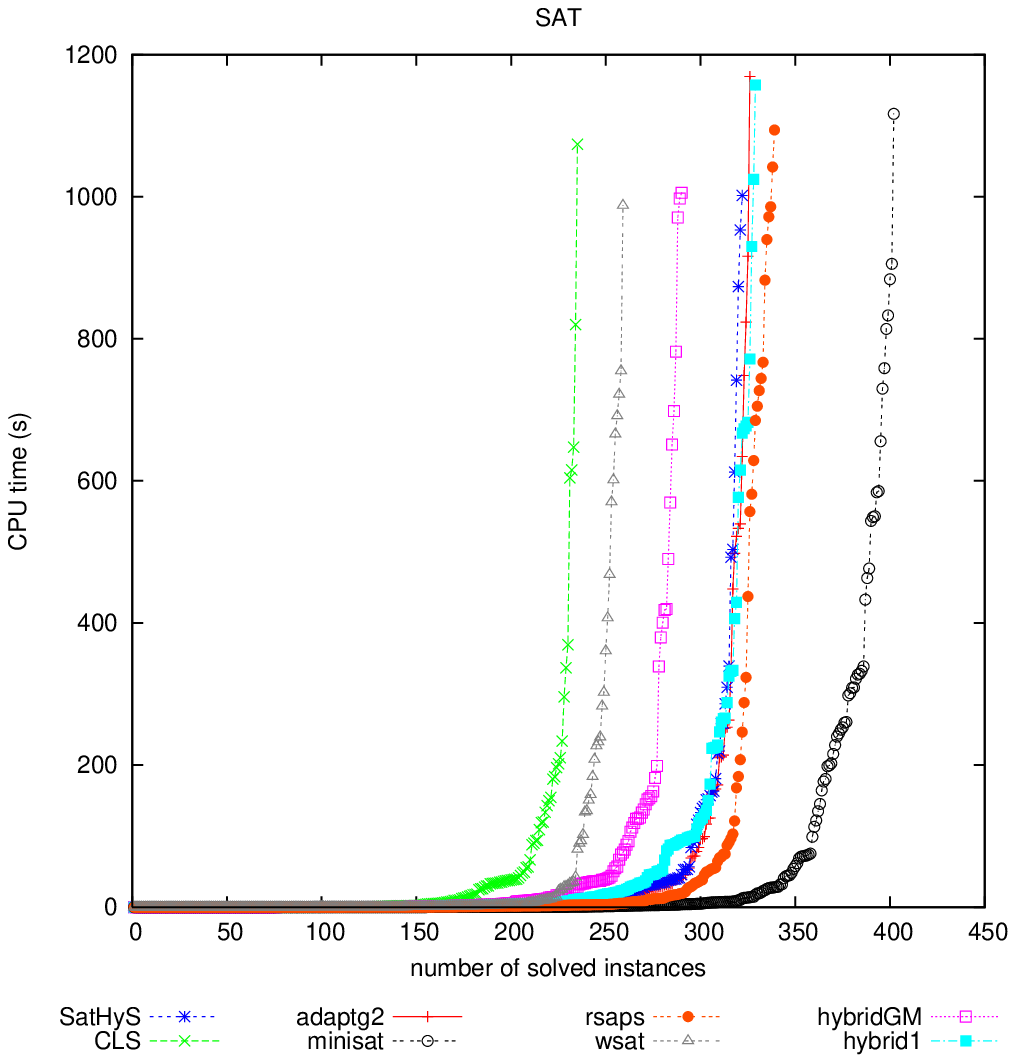,width=270pt}}
\caption{Crafted instances}\label{fig:crafted}
\end{figure}

\begin{figure}
      \centerline{\epsfig{file=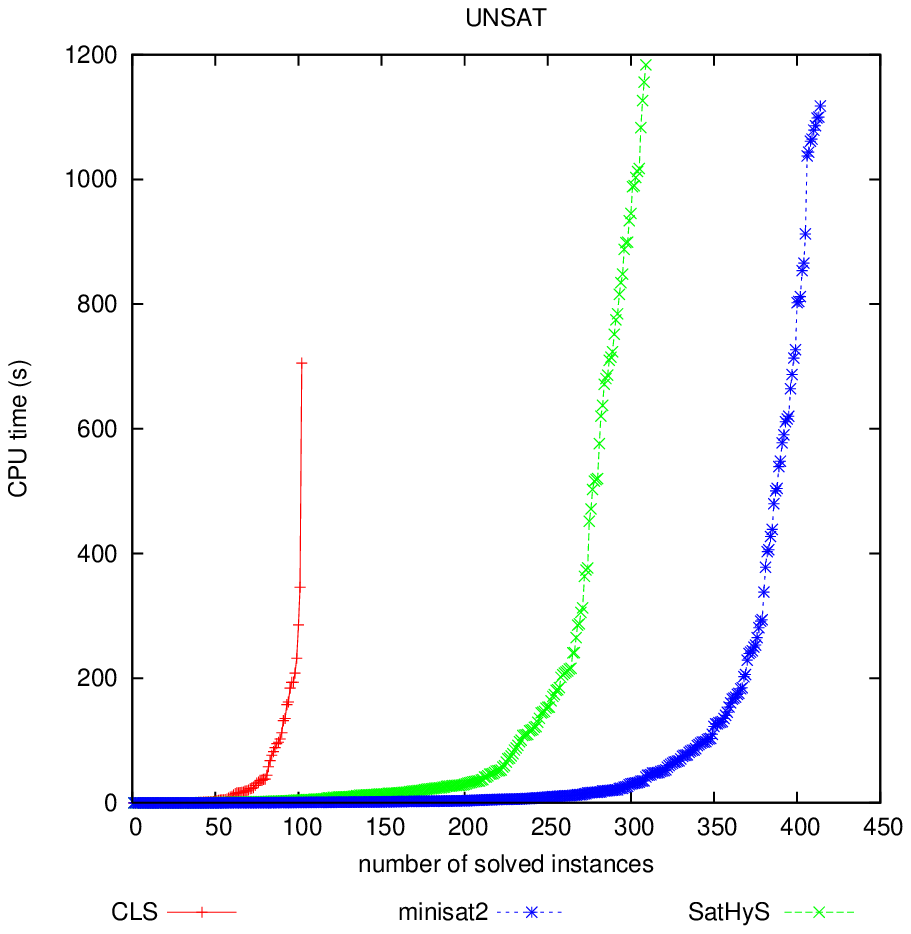,width=270pt}}
      \centerline{\epsfig{file=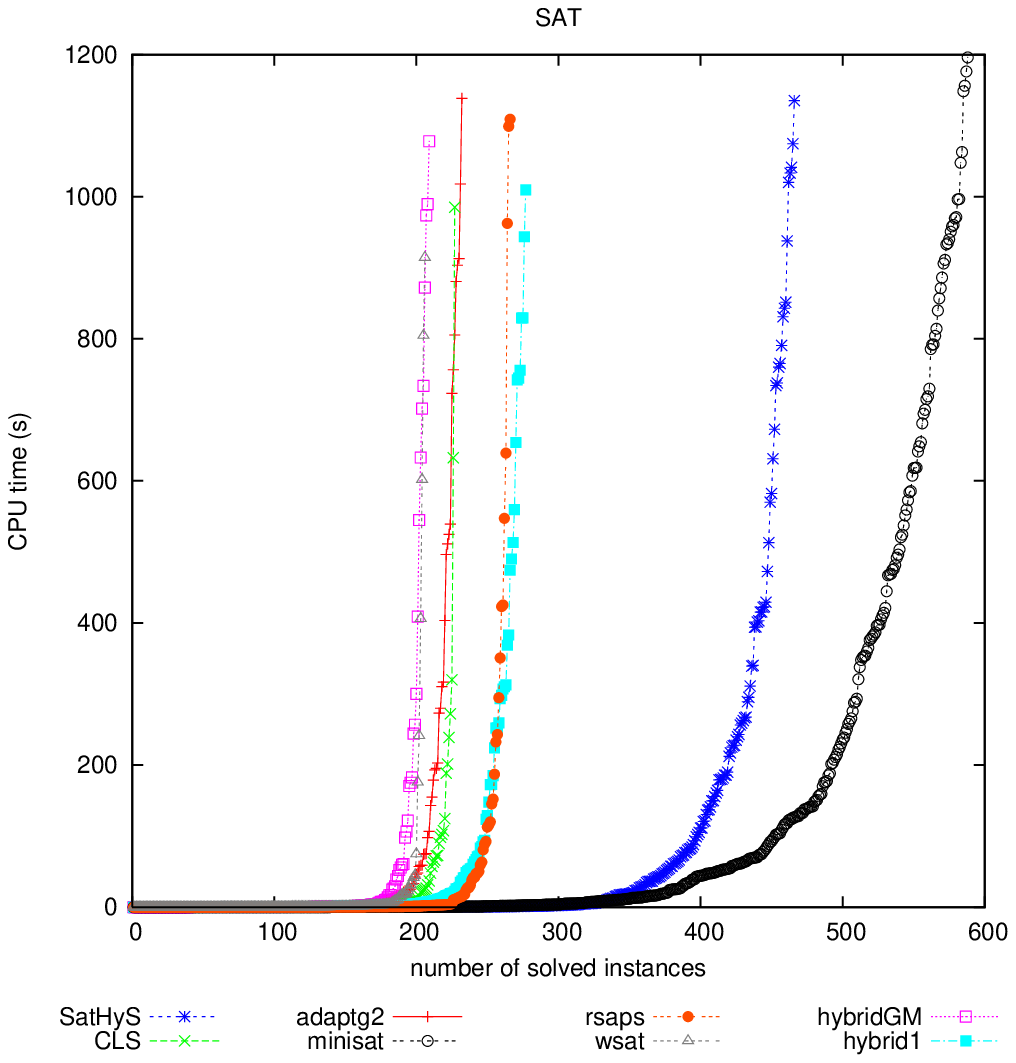,width=270pt}}
\caption{Industrial instances}\label{fig:indus}
\end{figure}

\begin{figure}
      \centerline{\epsfig{file=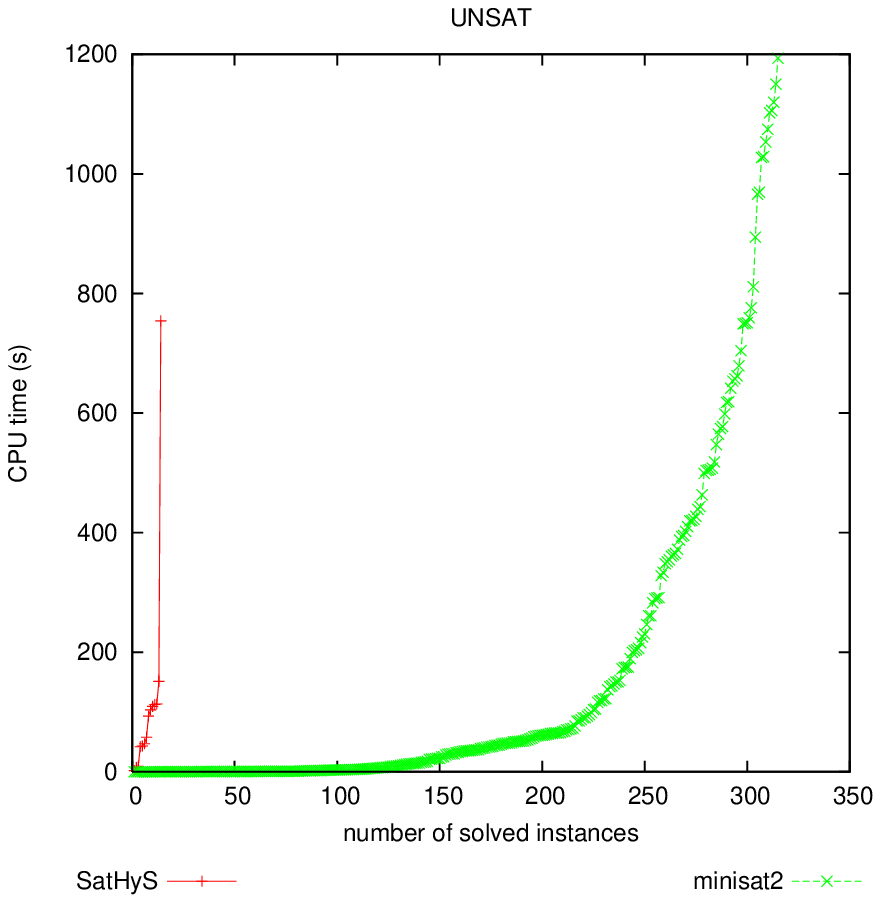,width=270pt}}
      \centerline{\epsfig{file=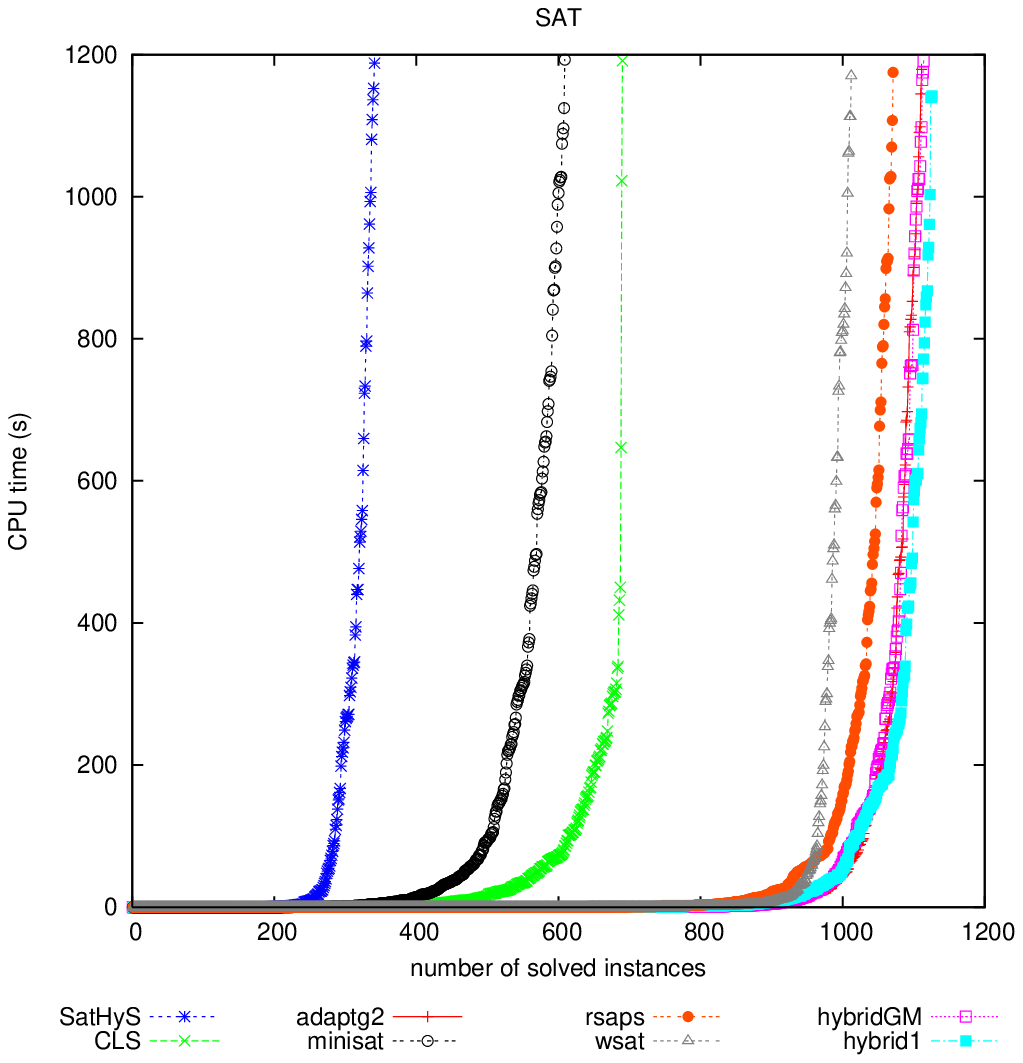,width=270pt}}
\caption{Random instances}\label{fig:random}
\end{figure}

The figures \ref{fig:crafted}, \ref{fig:indus} and \ref{fig:random}
give the classical cactus plot. For each tested method, the X-axis
corresponds to the number of formulas and the Y-axis corresponds to
the time needed to solve them if they were ran in parallel. When a
method does not appear in the curve, that means that this method is
not able to solve instance of this instances category. In these
figures, we have distinguished satisfiable and unsatisfiable instances
for each categories.

\section{Conclusion}
In this paper a new integration of local search and CDCL based SAT
solvers is introduced.  This hybrid solver represents an original
combination of both engines. The CDCL component can be seen as a new
strategy for escaping from local minimum.  This is achieved by the
assignment of opposite literals from the falsified clause.  In the
case of satisfiable SAT instances, such assignments are supposed to
behave like a tabu search approach, whereas for unsatisfiable ones,
they try to focus on a small sub-part of the formula, which is
minimally unsatisfiable (MUS).  \hybrid, the resulting method, obtains
very good results for a large category of instances. This new method
can be improved in different ways. As it was pointed in the
experimental section, our solver allows for more diversification and
less intensification.  First attempts have been done to correct
this. Finally, we aim at designing a solver which would focuses only
on an approximation of the MUS.

% This  paper presents  a new  hybridization of  solvers based  on local
% search  and  CDCL.  It  represents  an  original  combination of  both
% engines. The CDCL part of the solver is dedicated to escape from local
% minimum.  This is done by assignment of opposite literals of falsified
% clause.   In  case  of  satisfiable  problems,  such  assignments  are
% supposed to behave like a tabu search, whereas for unsatisfiable ones,
% they try to focus on a small sub-part of the formula, which is Minimum
% Unsatisfiable  Sub (formulas  (MUS).  \hybrid,  the  resulting method,
% obtains very good results for  a large category of instances. This new
% method can  be improved by  different ways. As  it was pointed  in the
% experimental section,  our solver allows for  more diversification and
% less  intensification.   First  attempts  have been  done  to  correct
% this. Furthermore, similarly as the behaviour of the CDCL component of
% our method  on unsatisfiable  problems, we aim  at designing  a solver
% which would focus only on MUS approximations.

\bibliographystyle{eptcs}
\bibliography{biblio}

\end{document}